\documentclass{article}
\usepackage[T1]{fontenc}
\usepackage{spconf,amsmath,amssymb,amsfonts,graphicx}
\usepackage{cite,textcomp,xcolor,url,booktabs,multirow}
\usepackage[hidelinks]{hyperref}

\title{ECG Mirage: Revealing and Mitigating the Underutilisation of ECGs in Vision-Language Models for Clinical Prediction}
\name{Jinning Liang \qquad Mingcheng Zhu \qquad Tingting Zhu}
\address{University of Oxford, Oxford, United Kingdom\\
jinning.liang@exeter.ox.ac.uk}

\begin{document}
\ninept
\maketitle

\begin{abstract}

Emergency department (ED) decision-making relies on heterogeneous clinical information, including patient history, vital signs, laboratory results, and electrocardiograms (ECGs). Vision--language models (VLMs) can jointly process these modalities, but strong predictive performance does not necessarily imply meaningful use of the correct patient's ECG. We term this failure mode ECG Mirage: apparent multimodal capability without useful dependence on patient-specific ECG information. We distinguish two forms: ECG neglect, where ECGs provide little predictive benefit, and ECG confusion, where matched ECGs outperform no-image inputs but not mismatched ECGs. To evaluate these behaviours, we compare predictions obtained with matched ECGs, outcome-discordant mismatched ECGs, and no-image inputs while holding the clinical text and prediction targets fixed. Across four VLMs on MDS-ED, matched ECGs provide no consistent advantage for either ICU admission or clinical deterioration prediction. We then train four restricted visual prompts using supervised learning followed by conditional direct preference optimisation, while keeping the VLM backbone frozen. The resulting models achieve balanced accuracies of 70.6\% for ICU admission and 67.5\% for deterioration and increase the matched-versus-mismatched performance gap to approximately 16.5 and 5.5 percentage points, respectively. Overall, our study identifies ECG Mirage in multimodal clinical prediction and introduces visual prompt tuning as an efficient mitigation strategy. Code is available at \url{https://github.com/JasonZuu/ECG-Mirage}.
\end{abstract}


\section{Introduction}
\label{sec:intro}

\begin{figure}[!t]
\centering
\includegraphics[
  width=\columnwidth,
  keepaspectratio
]{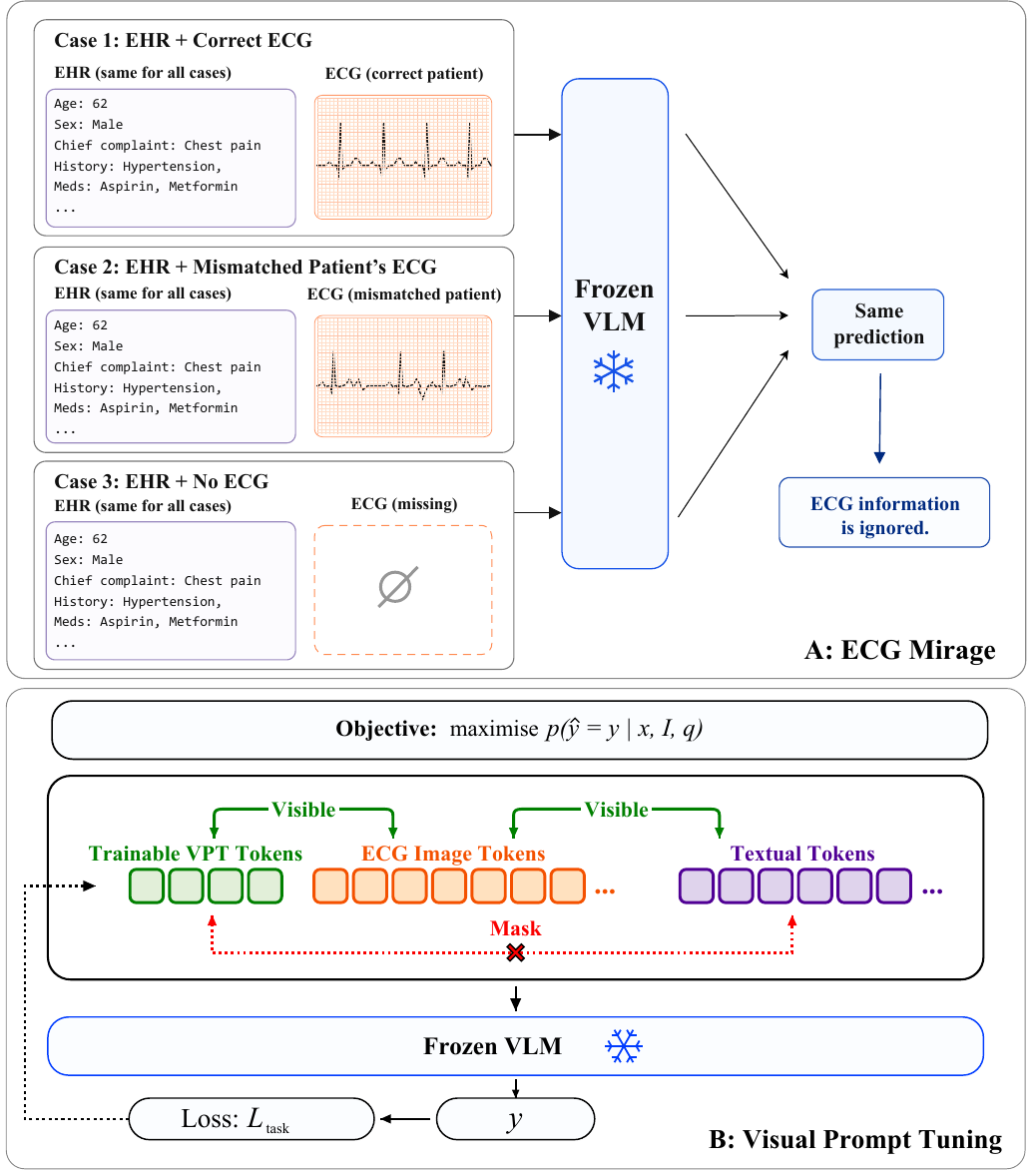}
\vspace{-10px}
\caption{Evaluation and training share a controlled ECG substitution. The reference answer remains fixed in conditional preference training. The no-image input contains only EHR text and the task question and is used only for evaluation.}
\label{fig:ecg-mirage}
\end{figure}

Predicting clinical outcomes in the emergency department (ED) presents an opportunity for multimodal signal processing from patient history, vital signs, laboratory measurements and electrocardiograms (ECGs)~\cite{chen2023multimodal,lopez2025mdsed}. Vision--language models (VLMs) offer a flexible approach to combining these sources, processing textual electronic health records (EHRs) alongside rendered ECG images~\cite{razzaq2026multimodal, zhu2026taxonomies}. Related EHR-based work has investigated efficient prompt compression for clinical prediction~\cite{zhu2026tokenpair} and adaptation across clinical conditions~\cite{zhu2026bridging}. Recent work has demonstrated progress in ECG interpretation and in linking diagnostic predictions to waveform evidence~\cite{liu2026teaching,lan2025gem}. These capabilities motivate the use of VLMs for outcome prediction, where ECGs must be interpreted alongside the broader clinical context. However, processing both modalities does not necessarily imply that both contribute to the prediction. This leads to a central question: do VLMs benefit from patient-specific ECG information beyond the EHR context?

Prior work shows that VLM benchmark performance can obscure limited visual dependence, including strong reliance on accompanying text in medical image prediction~\cite{chen2024we,buckley2026multimodal}. For clinical outcome prediction, an improvement from adding an ECG does not necessarily indicate that the model benefits from patient-specific information. A model could perform similarly when given another patient's ECG, despite outperforming a no-image control. We refer to this discrepancy as \emph{ECG Mirage}, where apparent multimodal predictive capability is not accompanied by reliable benefit from the ECG information. ECG Mirage operationalises this broader multimodal failure mode for clinical outcome prediction through patient-level ECG substitution controls. To examine this phenomenon, we compare matched ECGs, outcome-discordant mismatched ECGs and a no-image control while keeping the EHR, prediction question and target unchanged. Across the evaluated zero-shot VLMs and clinical tasks, matched ECGs offer no consistent predictive advantage over the controls. These findings motivate training that encourages models to distinguish matched from mismatched ECG evidence. 

To mitigate ECG Mirage, we propose visual prompt tuning (VPT), a lightweight training approach that combines prompt tuning with restricted visual prompts and conditional preference optimisation. The approach encourages VLMs to use patient-matched ECG information when predicting clinical outcomes from multimodal clinical data. Specifically, we train a small set of prompts while keeping the VLM backbone frozen. An attention mask prevents textual tokens from attending directly to these prompts, allowing them to influence predictions through only image tokens. We first train these prompts to predict clinical outcomes from matched EHR--ECG pairs. We then use conditional direct preference optimisation~\cite{dpo,wang2024mdpo} to encourage the model to favour the correct outcome when given the patient's own ECG over a mismatched ECG from another patient. Experiments on ICU admission and clinical deterioration prediction show that our approach mitigates ECG Mirage, improving matched-ECG balanced accuracy and widening the performance gap between matched- and mismatched-ECG cases. 

Our contributions are threefold. (1) We identify and characterise ECG Mirage across four zero-shot VLMs, showing that multimodal prediction performance does not necessarily reflect benefit from patient-matched ECGs. (2) We propose a lightweight approach that combines restricted visual prompt tuning with supervised learning and conditional preference optimisation to mitigate ECG Mirage. (3) Through experiments on two clinical prediction tasks, we demonstrate improvements in matched-ECG balanced accuracy and greater separation from mismatched ECGs and no-image inputs.

\section{Methodology}
\label{sec:methods}

\subsection{Clinical Prediction with VLMs}
\label{sec:prediction}

We consider clinical outcome prediction from electronic
health records (EHRs) and electrocardiogram (ECG) images. For each encounter, let $x$ denote the EHR input available at prediction time, $I$ the matched ECG image, and $q$ the task question. The target answer $y$ encodes either the binary ICU-admission outcome or the six clinical deterioration labels. We represent this answer as a token sequence $y=(y_1,\ldots,y_T)$. Given the EHR input $x$, ECG image $I$, and task question $q$, a VLM with parameters $\theta$ models
\begin{equation}
p_{\theta}(y\mid x,I,q) = \prod_{t=1}^{T} p_{\theta}(y_t\mid y_{<t},x,I,q).
\label{eq:prediction}
\end{equation}
We denote the resulting outcome prediction by $\hat{y}=f_{\theta}(x,I,q)$, using a fixed decoding procedure across input conditions.

\subsection{ECG Mirage}
\label{sec:ecg_mirage}

We define ECG Mirage as apparent multimodal predictive capability without reliable benefit from patient-matched ECG information. Let $I^{+}$ denote the matched ECG and $I^{-}$ an ECG from another patient with discordant outcomes. In the no-image condition ($\varnothing$), the VLM receives only the EHR input and task question. The EHR input $x$, task question $q$, and target $y$ remain fixed across conditions. For a task loss $\ell$, the expected prediction losses are
\begin{equation}
\begin{aligned}
R_{+} &= \mathbb{E}\!\left[\ell\!\left(f_{\theta}(x,I^{+},q),y\right)\right], \\
R_{-} &= \mathbb{E}\!\left[\ell\!\left(f_{\theta}(x,I^{-},q),y\right)\right], \\
R_{0} &= \mathbb{E}\!\left[\ell\!\left(f_{\theta}(x,\varnothing,q),y\right)\right].
\end{aligned}
\label{eq:ecg_risk}
\end{equation}

We assess whether patient-matched ECGs provide predictive benefit by requiring lower expected loss than both mismatched ECGs and EHR input alone:
\begin{equation}
R_{+} < R_{-} \quad \text{and} \quad R_{+} < R_{0}.
\label{eq:ecg_benefit}
\end{equation}

The first comparison assesses the benefit of the correct patient's ECG over an outcome-discordant substitute. The second assesses the added benefit of the ECG over the EHR text alone.

We distinguish two illustrative ECG Mirage patterns. ECG neglect occurs when $R_{+} \geq R_{-}$ and $R_{+} \geq R_{0}$, indicating no predictive advantage from the matched ECG over either control. ECG confusion occurs when $R_{-} \leq R_{+} < R_{0}$, indicating an advantage over the no-image condition without an advantage over the mismatched ECG. Thus, improvement from including an ECG does not necessarily establish benefit from the correct patient's ECG.
These conditions describe predictive behaviour rather than internal model processing. Poorer control performance can inflate the matched-ECG advantage without improving matched predictions.

\begin{figure*}[h!]
\centering
\includegraphics[width=0.98\textwidth]{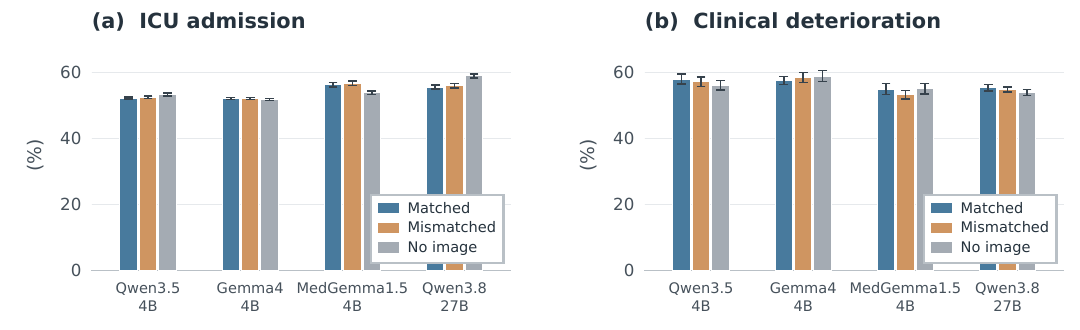}
\vspace{-15pt}
\caption{Zero-shot performance on ICU admission and clinical deterioration prediction under matched ECG, mismatched ECG and no-image conditions. Bars show full-test balanced accuracy for ICU admission and macro balanced accuracy for deterioration (\%). Error bars indicate one standard deviation from 100 case-bootstrap resamples.}
\label{fig:rq1}
\end{figure*}

\subsection{Visual Prompt Tuning (VPT) and Optimisation}
\label{sec:vpt_dpo}

We adapt visual prompt tuning~\cite{jia2022visual} through a restricted attention pathway and two-stage optimisation. 
Specifically, let $\mathbf{H}_{I}\in\mathbb{R}^{N_I\times d}$ denote the image token embeddings and $\mathbf{H}_{x,q}\in\mathbb{R}^{N_T\times d}$ the textual token embeddings, where $d$ is the hidden dimension. We introduce $K=4$ learnable visual prompt embeddings $\mathbf{H}_{v}\in\mathbb{R}^{K\times d}$ before the image and textual tokens, yielding
\begin{equation}
\mathbf{H}^{(0)} =
\bigl[\mathbf{H}_{v},\,\mathbf{H}_{I},\,\mathbf{H}_{x,q}\bigr],
\label{eq:visual_prompts}
\end{equation}
where the brackets denote concatenation along the token dimension. The backbone parameters $\theta$ remain frozen, and only $\mathbf{H}_{v}$ is optimised.
To restrict direct access to the prompts, let $\mathcal{P}$, $\mathcal{I}$, and $\mathcal{T}$ denote the positions of visual prompts, image tokens, and textual tokens, respectively. The textual positions include both input text and answer tokens. We define an additive attention mask
\begin{equation}
M_{ij} =
\begin{cases}
-\infty, & i\in\mathcal{T}\ \text{and}\ j\in\mathcal{P}, \\
0, & \text{otherwise},
\end{cases}
\label{eq:visual_mask}
\end{equation}
where $i$ and $j$ index queries and keys, respectively. For each attention head, we compute the masked attention as
\begin{equation}
\operatorname{Attention}(\mathbf{Q},\mathbf{K},\mathbf{V})
=
\operatorname{softmax}\!\left(
\frac{\mathbf{Q}\mathbf{K}^{\top}}{\sqrt{d_k}}
+\mathbf{A}+\mathbf{M}
\right)\mathbf{V},
\label{eq:masked_attention}
\end{equation}
where $\mathbf{Q}$, $\mathbf{K}$, and $\mathbf{V}$ are the query, key, and value matrices, $d_k$ is the key dimension, and $\mathbf{A}$ is the existing attention mask. In our implementation, image-token queries attend to preceding visual-prompt keys and values in full-attention layers, whereas textual queries are masked from attending directly to the prompts. The resulting prompt-conditioned image representations influence answer prediction through subsequent full-attention and recurrent linear-attention layers. To restrict direct prompt contributions through the recurrent pathway, we additionally zero the visual-prompt hidden states at the input to every recurrent token mixer of Qwen3.5. In the first training stage, we optimise the visual prompts on matched encounters. 
\begin{equation}
\mathcal{L}_{\mathrm{SFT}}
=
-\log p_{\theta,\mathbf{H}_v}(y\mid x,I^{+},q).
\label{eq:sft}
\end{equation}
Only target answer tokens contribute to this loss. This stage learns task-specific prompts through the restricted visual pathway, but provides no explicit comparison between matched and mismatched ECGs.
In the second stage, we initialise the policy and a frozen reference from the same supervised checkpoint. Each pair contains a matched ECG $I^{+}$ and a mismatched ECG $I^{-}$, with the EHR input $x$, task question $q$, and target answer $y$ held fixed. We define the policy's corresponding log-probability margin as
\begin{equation}
m_{\mathbf{H}_v}
=
\log p_{\theta,\mathbf{H}_v}(y\mid x,I^{+},q)
-
\log p_{\theta,\mathbf{H}_v}(y\mid x,I^{-},q).
\label{eq:preference_margin}
\end{equation}

Let $m_{\mathrm{ref}}$ denote the same margin under the frozen reference. The second-stage training objective is
\begin{equation}
\mathcal{L}_{\mathrm{DPO}}
=
-\log \sigma\!\left(\beta[m_{\mathbf{H}_v}-m_{\mathrm{ref}}]\right)
+
\lambda\mathcal{L}_{\mathrm{SFT}},
\label{eq:dpo}
\end{equation}
where $\sigma$ is the logistic sigmoid, $\beta=0.1$ scales the preference margin, and $\lambda=0.1$ weights the supervised term. DPO favours the target answer under matched over mismatched ECGs.

\section{Results}
\label{sec:results}
We investigate whether zero-shot VLMs benefit from patient-matched ECGs in clinical outcome prediction (RQ1), whether our VPT approach improves predictive performance while mitigating ECG Mirage (RQ2), and how conditional preference optimisation, the supervised training stage and the visual attention mask contribute to these outcomes (RQ3).


\subsection{Experimental setup}
\label{sec:experimental_setup}

We use MIMIC-IV-ED~\cite{johnson2023mimiced} with linked MIMIC-IV~\cite{johnson2023mimiciv} and MIMIC-IV-ECG ~\cite{gow2023mimicivecg} following MDS-ED cohort construction and patient-level partitioning~\cite{lopez2025mdsed}. 
For ICU admission, the training, validation and test sets comprise 108,877/\allowbreak 5,802/\allowbreak 6,048 visits from 63,676/\allowbreak 3,513/\allowbreak 3,626 patients, respectively. For deterioration, the corresponding sets comprise 109,299/\allowbreak 5,819/\allowbreak 6,077 visits from 63,929/\allowbreak 3,525/\allowbreak 3,644 patients.
ICU admission is assessed over the entire hospital stay, while deterioration outcomes are assessed within 24 hours of ED arrival, using the records from the first 90 minutes of observation in the ED admission~\cite{lopez2025mdsed}. The clinical deterioration task comprises six outcomes: severe hypoxaemia, vasopressor use, mechanical ventilation, extracorporeal membrane oxygenation, inotrope use, and in-hospital cardiac arrest.

Prediction is performed 90 minutes after ED arrival. All EHR information available within this window is converted into textual input following~\cite{chen2026cross}, with task-specific leakage exclusions applied. The matched ECG is the first recording from the ED encounter acquired by prediction time. We compare matched, mismatched, and no-image conditions while keeping the EHR input, task question, and target fixed. The mismatched condition randomly selects an ECG from another patient with a different task label. The no-image condition supplies only EHR text and the task question.

For the deep learning baseline, we use the S4--MLP model~\cite{lopez2025mdsed}, which combines an ECG waveform encoder with an MLP for fusion with tabular clinical features. For VLM adaptation baselines, we include low-rank adaptation (LoRA)~\cite{hu2022lora} and prompt tuning~\cite{lester2021prompt}. All methods use the same cohort splits, prediction-time cutoff, and leakage exclusions. Models are trained separately for the two prediction tasks.
For S4--MLP, we follow the original training configuration reported in MDS-ED~\cite{lopez2025mdsed}. For VLM adaptation, we use AdamW~\cite{loshchilov2019adamw} with a learning rate of $10^{-3}$ and a batch size of 32. The learning rate increases linearly over the first 10\% of training steps and follows a cosine decay schedule over the remaining 90\%. Training runs for a maximum of one epoch, with validation every 500 optimisation steps. We select the best checkpoint by task-specific validation F1 and stop after three consecutive checks without improvement.

For ICU admission, we report balanced accuracy and macro F1. For clinical deterioration, we report macro balanced accuracy and macro positive-class F1 across the six outcomes, excluding unavailable labels from the corresponding calculations. We report scores computed on the complete test set together with standard deviations estimated from 100 bootstrap resamples~\cite{efron1979bootstrap}. Each resample draws test encounters with replacement, retaining their predictions and reference labels, and the same resampling indices are used across methods and input conditions. These standard deviations describe test-set sampling variability rather than variation across training runs.

\begin{table*}[t]
\centering
\small
\setlength{\tabcolsep}{4pt}
\renewcommand{\arraystretch}{1.0}
\caption{Performance on ICU admission and clinical deterioration. All VLM methods use Qwen3.5-4B. S4-MLP follows the MDS-ED architecture and main training hyperparameters. Subscripts indicate case-bootstrap standard deviations over 100 resamples.}
\vspace{4px}
\label{tab:rq2}
\resizebox{\textwidth}{!}{%
\begin{tabular}{@{}l*{12}{c}@{}}
\toprule
& \multicolumn{6}{c}{ICU Admission} & \multicolumn{6}{c}{Clinical Deterioration} \\
\cmidrule(lr){2-7}\cmidrule(lr){8-13}
& \multicolumn{2}{c}{Matched} & \multicolumn{2}{c}{Mismatched} & \multicolumn{2}{c}{No image}
& \multicolumn{2}{c}{Matched} & \multicolumn{2}{c}{Mismatched} & \multicolumn{2}{c}{No image} \\
\cmidrule(lr){2-3}\cmidrule(lr){4-5}\cmidrule(lr){6-7}\cmidrule(lr){8-9}\cmidrule(lr){10-11}\cmidrule(lr){12-13}
Method & BAcc & F1 & BAcc & F1 & BAcc & F1 & BAcc & F1 & BAcc & F1 & BAcc & F1 \\
\midrule
S4-MLP & $67.3_{\pm0.8}$ & $49.6_{\pm1.6}$ & --- & --- & --- & --- & $55.0_{\pm0.8}$ & $12.8_{\pm1.7}$ & --- & --- & --- & --- \\
\midrule
Zero-shot & $52.3_{\pm0.3}$ & $50.7_{\pm0.7}$ & $52.5_{\pm0.4}$ & $51.2_{\pm0.7}$ & $53.4_{\pm0.4}$ & $52.9_{\pm0.8}$ & $58.1_{\pm1.6}$ & $12.8_{\pm2.0}$ & $57.2_{\pm1.5}$ & $11.8_{\pm2.0}$ & $56.2_{\pm1.5}$ & $13.1_{\pm2.8}$ \\
LoRA & $74.7_{\pm1.0}$ & $78.8_{\pm0.9}$ & $73.8_{\pm1.0}$ & $78.0_{\pm0.9}$ & $74.7_{\pm1.0}$ & $78.5_{\pm0.9}$ & $69.2_{\pm2.4}$ & $34.4_{\pm3.3}$ & $68.4_{\pm2.3}$ & $33.5_{\pm3.2}$ & $68.4_{\pm2.4}$ & $30.3_{\pm2.4}$ \\
Prompt Tuning & $71.2_{\pm0.9}$ & $75.4_{\pm0.9}$ & $70.6_{\pm0.9}$ & $74.7_{\pm0.9}$ & $54.3_{\pm0.4}$ & $54.3_{\pm0.8}$ & $67.1_{\pm2.0}$ & $22.8_{\pm2.3}$ & $66.7_{\pm2.1}$ & $23.0_{\pm2.3}$ & $63.8_{\pm1.6}$ & $17.2_{\pm2.0}$ \\
\midrule
VPT (Ours) & $70.6_{\pm0.9}$ & $62.8_{\pm0.7}$ & $54.1_{\pm1.0}$ & $47.7_{\pm0.6}$ & $55.5_{\pm0.5}$ & $56.5_{\pm0.9}$ & $67.5_{\pm2.1}$ & $21.7_{\pm2.8}$ & $62.1_{\pm1.6}$ & $17.4_{\pm1.7}$ & $55.6_{\pm1.2}$ & $11.0_{\pm1.9}$ \\
\bottomrule
\end{tabular}%
}
\end{table*}

\subsection{RQ1: Examining ECG Mirage in zero-shot prediction}
In this study, we examined whether zero-shot VLMs derive predictive benefit from ECGs when EHR is also available. We evaluated Qwen3.5-4B~\cite{qwen2026qwen35}, Qwen3.8-27B~\cite{qwen2026qwen35}, Gemma4-4B~\cite{team2026gemma} and MedGemma1.5-4B~\cite{sellergren2026medgemma} on ICU admission and clinical deterioration prediction under three conditions: matched ECGs, mismatched ECGs from other patients, and no image. 

Figure~\ref{fig:rq1} shows evidence of ECG Mirage. For ICU admission, all four models had slightly lower balanced-accuracy point estimates with matched ECGs than with mismatched ECGs. Qwen3.5-4B achieved 52.3\%, 52.5\% and 53.4\% under matched, mismatched and no-image conditions, respectively. MedGemma performed better with matched ECGs than without an image, but not better than with mismatched ECGs, indicating that gains from including an ECG do not necessarily reflect a benefit from patient-specific information. For clinical deterioration, Gemma4-4B also performed worse with matched ECGs than under either control, whereas both Qwen models achieved higher matched-ECG point estimates than under either control. Overall, matched ECGs provided no consistent predictive advantage across the evaluated models and tasks, supporting the existence of ECG Mirage. These models may rely primarily on EHR text, consistent with prior findings that medical images add little predictive value when clinical text is informative~\cite{buckley2026multimodal}.

\subsection{RQ2: Mitigating ECG Mirage}
We investigated whether our VPT approach mitigates ECG Mirage while improving clinical outcome prediction. Table~\ref{tab:rq2} compares VPT with zero-shot inference, LoRA and supervised prompt tuning, using Qwen3.5-4B as the VLM backbone. S4--MLP served as a deep-learning baseline for predictive performance. 

VPT improved matched-ECG performance over zero-shot inference on both tasks, achieving balanced accuracies of 70.6\% for ICU admission and 67.5\% for clinical deterioration, with corresponding F1 scores of 62.8\% and 21.7\% under the task-specific definitions. Matched ECGs outperformed mismatched ECGs by 16.5 and 5.5 percentage points, respectively, and no-image inputs by 15.1 and 11.9 percentage points. VPT had the largest matched–mismatched balanced-accuracy gaps among the evaluated VLM methods. Although LoRA achieved higher absolute predictive performance, its performance remained similar across ECG conditions.
These comparisons show that improvements in task performance need not be accompanied by greater benefit from patient-matched ECGs. LoRA substantially improved prediction even without an image, while the prompt-tuning results indicate that a benefit from adding an ECG need not depend on patient matching. Our VPT approach combined improved matched performance over zero-shot inference with higher scores for matched ECGs than for either control, supporting mitigation of ECG Mirage in the evaluated setting. Its main benefit was therefore improved prediction over the unadapted model combined with a clearer distinction between matched and mismatched ECGs.

\begin{table}[h!]
\centering
\small
\setlength{\tabcolsep}{3pt}
\renewcommand{\arraystretch}{1.0}
\caption{Ablation study on clinical deterioration prediction. Performance is measured with balanced accuracy (\%). }
\vspace{2pt}
\label{tab:rq3}
\begin{tabular}{@{}lccc@{}}
\toprule
Method & Matched & Mismatched & No image \\
\midrule
VPT & $67.5_{\pm2.1}$ & $62.1_{\pm1.6}$ & $55.6_{\pm1.2}$ \\
\midrule
w/o DPO & $63.4_{\pm1.6}$ & $63.7_{\pm1.8}$ & $55.4_{\pm1.1}$ \\
w/o SFT Stage & $56.5_{\pm1.4}$ & $56.3_{\pm1.4}$ & $55.5_{\pm1.2}$ \\
w/o Visual Mask & $64.3_{\pm1.8}$ & $64.1_{\pm1.7}$ & $62.2_{\pm1.6}$ \\
\bottomrule
\end{tabular}
\end{table}

\subsection{RQ3: Contributions of the adaptation components}

We conducted ablation experiments on clinical deterioration prediction to assess the contributions of conditional preference optimisation, supervised initialisation and the visual attention mask. Table~\ref{tab:rq3} compares VPT with three variants: w/o DPO, which uses supervised training alone; w/o SFT stage, which removes the separate supervised training stage and jointly optimises the SFT and DPO losses from the outset; and w/o Visual Mask, which allows text tokens to attend directly to the learned prompts. 
VPT achieved the highest matched balanced accuracy among the evaluated configurations (67.5\%). The w/o DPO variant achieved similar scores with matched (63.4\%) and mismatched (63.7\%) ECGs, although both exceeded its no-image performance (55.4\%). Relative to this variant, VPT improved matched performance and reduced mismatched performance to 62.1\%, with both changes contributing to the larger separation. The Joint SFT+DPO variant achieved 56.5\% with matched ECGs and similar scores under mismatched (56.3\%) and no-image (55.5\%) conditions, favouring a separate supervised initialisation stage over joint optimisation from the outset in this setting. The w/o Visual Mask variant retained sequential training but achieved similar matched (64.3\%) and mismatched (64.1\%) scores. 

\section{Conclusion}
\label{sec:conclusion}
In this study, we identified ECG Mirage in multimodal clinical outcome prediction, where VLM performance can obscure benefit from patient-matched ECG information. Across four zero-shot VLMs and two ED prediction tasks, matched ECGs offered no consistent advantage over mismatched ECGs or EHR text alone. We introduced a lightweight approach combining restricted visual prompts and conditional preference optimisation with a frozen backbone. VPT improved matched-ECG performance over zero-shot inference and increased matched–mismatched performance gaps on both tasks. These findings support evaluating multimodal clinical models through both predictive performance and patient-matched modality benefit. However, LoRA achieved higher predictive performance with matched ECGs than VPT on both tasks. Future work aims to improve predictive performance while ensuring that models benefit from patient-specific ECG information.

\section{Compliance with Ethical Standards}
This study retrospectively analysed de-identified data from MIMIC-IV, accessed through PhysioNet under the respective data use agreements. The Beth Israel Deaconess Medical Centre Institutional Review Board approved the original MIMIC-IV data sharing with a waiver of informed consent. Our study involved no patient recruitment or clinical intervention. 

\section{Acknowledgements}
The authors declare no conflicts of interest.

\bibliographystyle{IEEEbib}
\bibliography{refs}

@inproceedings{dpo,
  author = {Rafailov, Rafael and Sharma, Archit and Mitchell, Eric and Manning, Christopher D. and Ermon, Stefano and Finn, Chelsea},
  title = {Direct Preference Optimization: Your Language Model Is Secretly a Reward Model},
  booktitle = {Advances in Neural Information Processing Systems},
  volume = {36},
  pages = {53728--53741},
  year = {2023}
}

@article{johnson2023mimiced,
  title = {{MIMIC-IV-ED}},
  author = {Johnson, Alistair and Bulgarelli, Lucas and Pollard, Tom and Celi, Leo Anthony and Mark, Roger and Horng, Steven},
  journal = {PhysioNet},
  year = {2023},
  note = {Version 2.2},
  doi = {10.13026/5ntk-km72},
  url = {https://physionet.org/content/mimic-iv-ed/2.2/}
}

@article{johnson2023mimiciv,
  title = {{MIMIC-IV}},
  author = {Johnson, Alistair and Bulgarelli, Lucas and Pollard, Tom and Horng, Steven and Celi, Leo Anthony and Mark, Roger},
  journal = {PhysioNet},
  year = {2023},
  note = {Version 2.2},
  doi = {10.13026/6mm1-ek67},
  url = {https://physionet.org/content/mimiciv/2.2/}
}

@article{lopez2025mdsed,
  title = {Enhancing clinical decision support with physiological waveforms---A multimodal benchmark in emergency care},
  author = {{Lopez Alcaraz}, Juan Miguel and Bouma, Hjalmar and Strodthoff, Nils},
  journal = {Computers in Biology and Medicine},
  volume = {192},
  pages = {110196},
  year = {2025},
  doi = {10.1016/j.compbiomed.2025.110196}
}

@article{zhu2026taxonomies,
  title={The Taxonomies, Training, and Applications of Event Stream Modelling for Electronic Health Records},
  author={Zhu, Mingcheng and Liu, Yu and Luo, Zhiyao and Zhu, Tingting},
  journal={arXiv preprint arXiv:2603.14003},
  year={2026}
}

@inproceedings{chen2026cross,
  title={Cross-representation benchmarking in time-series electronic health records for clinical outcome prediction},
  author={Chen, Tianyi and Zhu, Mingcheng and Luo, Zhiyao and Zhu, Tingting},
  booktitle={ICASSP 2026-2026 IEEE International Conference on Acoustics, Speech and Signal Processing (ICASSP)},
  pages={7076--7080},
  year={2026},
  organization={IEEE}
}

@inproceedings{hu2022lora,
  title = {{LoRA}: Low-Rank Adaptation of Large Language Models},
  author = {Hu, Edward J. and Shen, Yelong and Wallis, Phillip and Allen-Zhu, Zeyuan and Li, Yuanzhi and Wang, Shean and Wang, Lu and Chen, Weizhu},
  booktitle = {International Conference on Learning Representations},
  year = {2022},
  url = {https://openreview.net/forum?id=nZeVKeeFYf9}
}

@inproceedings{lester2021prompt,
  title = {The Power of Scale for Parameter-Efficient Prompt Tuning},
  author = {Lester, Brian and Al-Rfou, Rami and Constant, Noah},
  booktitle = {Proceedings of the 2021 Conference on Empirical Methods in Natural Language Processing},
  pages = {3045--3059},
  year = {2021},
  publisher = {Association for Computational Linguistics},
  doi = {10.18653/v1/2021.emnlp-main.243},
  url = {https://aclanthology.org/2021.emnlp-main.243/}
}

@inproceedings{loshchilov2019adamw,
  title = {Decoupled Weight Decay Regularization},
  author = {Loshchilov, Ilya and Hutter, Frank},
  booktitle = {International Conference on Learning Representations},
  year = {2019},
  url = {https://openreview.net/forum?id=Bkg6RiCqY7}
}

@article{efron1979bootstrap,
  title = {Bootstrap Methods: Another Look at the Jackknife},
  author = {Efron, Bradley},
  journal = {The Annals of Statistics},
  volume = {7},
  number = {1},
  pages = {1--26},
  year = {1979},
  publisher = {Institute of Mathematical Statistics},
  doi = {10.1214/aos/1176344552}
}

@article{chen2023multimodal,
  title={Multimodal clinical benchmark for emergency care ({MC-BEC}): A comprehensive benchmark for evaluating foundation models in emergency medicine},
  author={Chen, Emma and others},
  journal={Advances in Neural Information Processing Systems},
  volume={36},
  pages={45794--45811},
  year={2023}
}

@article{liu2026teaching,
  title={Teaching multimodal {LLMs} to comprehend 12-lead electrocardiographic images},
  author={Liu, Ruoqi and Bai, Yuelin and Yue, Xiang and Zhang, Ping},
  journal={npj Digital Medicine},
  volume={9},
  number={1},
  pages={349},
  year={2026},
  publisher={Nature Publishing Group UK London}
}

@article{lan2025gem,
  title={{GEM}: Empowering {MLLM} for grounded {ECG} understanding with time series and images},
  author={Lan, Xiang and Wu, Feng and He, Kai and Zhao, Qinghao and Hong, Shenda and Feng, Mengling},
  journal={Advances in Neural Information Processing Systems},
  volume={38},
  pages={94421--94455},
  year={2025}
}

@article{buckley2026multimodal,
  title={Multimodal foundation models exploit text to make medical image predictions},
  author={Buckley, Thomas A and Diao, James A and Srivastava, Cam N and Brodeur, Peter G and Rajpurkar, Pranav and Rodman, Adam and Manrai, Arjun K},
  journal={Nature Communications},
  year={2026},
  publisher={Nature Publishing Group},
  volume={17},
  pages={7475}
}

@article{chen2024we,
  title={Are we on the right way for evaluating large vision-language models?},
  author={Chen, Lin and Li, Jinsong and Dong, Xiaoyi and Zhang, Pan and Zang, Yuhang and Chen, Zehui and Duan, Haodong and Wang, Jiaqi and Qiao, Yu and Lin, Dahua and others},
  journal={Advances in Neural Information Processing Systems},
  volume={37},
  pages={27056--27087},
  year={2024}
}

@inproceedings{wang2024mdpo,
  title={{mDPO}: Conditional preference optimization for multimodal large language models},
  author={Wang, Fei and Zhou, Wenxuan and Huang, James Y and Xu, Nan and Zhang, Sheng and Poon, Hoifung and Chen, Muhao},
  booktitle={Proceedings of the 2024 Conference on Empirical Methods in Natural Language Processing},
  pages={8078--8088},
  year={2024}
}

@misc{qwen2026qwen35,
  title={{Qwen3.5}: Towards Native Multimodal Agents},
  author={{Qwen Team}},
  year={2026},
  howpublished={\url{https://qwen.ai/blog?id=qwen3.5}}
}

@article{gow2023mimicivecg,
  title={{MIMIC-IV-ECG}: Diagnostic electrocardiogram matched subset},
  author={Gow, Brian and Pollard, Tom and Nathanson, Larry A and Johnson, Alistair and Moody, Benjamin and Fernandes, Chrystinne and Greenbaum, Nathaniel and Waks, Jonathan W and Eslami, Parastou and Carbonati, Tanner and others},
  journal={PhysioNet},
  year={2023},
  note={Version 1.0},
  doi={10.13026/4nqg-sb35}
}

@inproceedings{jia2022visual,
  title={Visual prompt tuning},
  author={Jia, Menglin and Tang, Luming and Chen, Bor-Chun and Cardie, Claire and Belongie, Serge and Hariharan, Bharath and Lim, Ser-Nam},
  booktitle={European conference on computer vision},
  pages={709--727},
  year={2022},
  organization={Springer}
}

@inproceedings{zhu2026tokenpair,
  title={From Token to Token Pair: Efficient Prompt Compression for Large Language Models in Clinical Prediction},
  author={Zhu, Mingcheng and Luo, Zhiyao and Liu, Yu and Zhu, Tingting},
  booktitle={Proceedings of the 43rd International Conference on Machine Learning},
  year={2026}
}

@article{zhu2026bridging,
  title={Bridging data gaps of rare conditions in {ICU}: a multi-disease adaptation approach for clinical prediction},
  author={Zhu, Mingcheng and Liu, Yu and Luo, Zhiyao and Zhu, Tingting},
  journal={npj Digital Medicine},
  volume={9},
  number={1},
  pages={7},
  year={2026},
  publisher={Nature Publishing Group UK London}
}

@article{team2026gemma,
  title={{Gemma 4} technical report},
  author={Team, Gemma and Abd, Sherif El and Aggarwal, Vaibhav and Algayres, Robin and Andreev, Alek and Bachem, Olivier and Ballantyne, Ian and Brick, Cormac and C{\u{a}}rbune, Victor and Casbon, Michelle and others},
  journal={arXiv preprint arXiv:2607.02770},
  year={2026}
}

@article{sellergren2026medgemma,
  title={{MedGemma 1.5} technical report},
  author={Sellergren, Andrew and others},
  journal={arXiv preprint arXiv:2604.05081},
  year={2026}
}

@article{razzaq2026multimodal,
  title={Multimodal AI in healthcare: Review of vision-language foundation models for real-world medical applications},
  author={Razzaq, Taha and Taj, Murtaza and Iqbal, Asim},
  journal={Journal of Biomedical Informatics},
  pages={105075},
  year={2026},
  publisher={Elsevier}
}
\end{document}